%% file: ijcai23.tex
\documentclass{article}
\usepackage{ijcai23}

\usepackage{times}
\usepackage{soul}
\usepackage{url}
\usepackage[hidelinks]{hyperref}
\usepackage[utf8]{inputenc}
\usepackage[small]{caption}
\usepackage{graphicx}
\usepackage{amsmath}
\usepackage{amsthm}
\usepackage{booktabs}
\usepackage{algorithm}
\usepackage{algorithmic}
\usepackage{dirtytalk}
\usepackage[switch]{lineno}

\usepackage{latexsym}

\usepackage{subcaption}

\usepackage{algorithm}
\usepackage{bbm}
\usepackage{algorithmic}

\newcommand{\name}{Deep Symbolic Learning }
\newcommand{\shortname}{DSL}
\newcommand{\godel}{G\"{o}del}
\newcommand{\digit}[1]{\vcenter{\hbox{\includegraphics[height=10pt]{#1}}}}

\newcommand{\Sym}{\mathcal{S} }
\newcommand{\G}{{\bf G}}
\newcommand{\W}{{\bf W}}

\usepackage{newfloat}
\usepackage{listings}
\usepackage{amsfonts,amsmath}
\usepackage{bm}

\DeclareMathOperator*{\argmin}{argmin}
\usepackage{tikz}
\usetikzlibrary{positioning,decorations.pathreplacing,shapes}
\newtheorem{definition}{Definition}
\newtheorem{example}{Example}

\title{
Deep Symbolic Learning: Discovering Symbols and Rules 
from Perceptions
}

\author{
    Alessandro Daniele\textsuperscript{\rm 1}
    \and
    Tommaso Campari\textsuperscript{\rm 1, \rm 2}
    \and
    Sagar Malhotra\textsuperscript{\rm 1, \rm 3}\And
    Luciano Serafini\textsuperscript{\rm 1}
\affiliations
\textsuperscript{\rm 1}Fondazione Bruno Kessler, Trento, Italy \\
    \textsuperscript{\rm 2}Universit\`a degli Studi di Padova, Italy\\
    \textsuperscript{\rm 3}Universit\`a degli Studi di Trento, Italy 
    \emails
    \{daniele, tcampari, smalhotra, serafini\}@fbk.eu
}

\begin{document}

\maketitle

\maketitle
\input{chapters/abstract}
\section{Introduction}
\input{chapters/introduction}

\section{Related Works}
\input{chapters/related_work}

\input{chapters/problem_definition}

\section{Method}
\input{chapters/method}

\section{Experiments}
\label{sec:experiments}
\input{chapters/experiments}

\section{Conclusion and Future Work}
\input{chapters/conclusion}

\bibliographystyle{named}
\bibliography{ijcai23}
\clearpage
\newpage
\appendix

\input{chapters/appendix.tex}


\end{document}

%% file: chapters/abstract.tex
\begin{abstract}

  Neuro-Symbolic (NeSy) integration combines symbolic reasoning with Neural Networks (NNs) for tasks requiring perception and reasoning. Most NeSy systems rely on continuous relaxation of logical knowledge, and no discrete decisions are made within the model pipeline. Furthermore, these methods assume that the symbolic rules are given. In this paper, we propose \emph{\name} (\shortname), a NeSy system that learns \emph{NeSy-functions}, i.e., the composition of a (set of) \emph{perception functions} which map continuous data to discrete symbols, and a \emph{symbolic function} over the set of symbols. \shortname\ simultaneously learns the perception and symbolic functions while being trained only on their composition (NeSy-function). The key novelty of \shortname\ is that it can create internal (interpretable) symbolic representations and map them to perception inputs within a differentiable NN learning pipeline. The created symbols are automatically selected to generate symbolic functions that best explain the data. We provide experimental analysis to substantiate the efficacy of \shortname\  in simultaneously learning perception and symbolic functions.
  %
\end{abstract}

%% file: chapters/introduction.tex
Neuro-Symbolic (NeSy) Systems combine deep neural networks and symbolic reasoning so that learning and reasoning can occur in a symbiotic fashion. The fundamental goal of NeSy systems is to incorporate and potentially learn the symbolic rules while still exploiting neural networks (NNs) for interpreting perception and guiding exploration in the combinatorial search space. In general, a NeSy system can be seen as a composition of \emph{perception functions} and \emph{symbolic functions}. Perception functions map perception, usually represented as real-valued tensors, to symbols, whereas symbolic functions map symbols to other symbols. The first challenge to any NeSy system is to reconcile the dichotomy between the intrinsically discrete nature of symbolic reasoning and the implicit continuity requirement of gradient descent-based learning methods. Recent works have tried to resolve this problem by exploiting different types of continuous relaxations to logical rules. However, with few exceptions, most such works assume the symbolic functions to be given a priori, and they use these functions to guide the training of a perception function, parameterized as a NN. A key challenge to such systems is the lack of a method capable of performing symbolic manipulations and meaningfully associating symbols to perception inputs, also known as the \emph{Symbol Grounding Problem} \cite{Symbol_Grounding_Problem}.

In this paper, we introduce the concept of \emph{NeSy-function}, i.e., a composition of a set of perception and symbolic functions.
Moreover, we propose \emph{\name (\shortname)}, a framework that can jointly learn perception and symbolic functions while supervised only on the NeSy function. This is done by introducing policy functions, similar to Reinforcement Learning (RL) \cite{RL}, within the neural architecture. The policy function chooses internal symbolic representations to be associated with the perception inputs based on the confidence values generated by the neural networks. The selected symbols are then combined to form a unique prediction for the NeSy function, while their confidences are interpreted under fuzzy logic semantics to estimate the confidence of such a prediction. Moreover, \shortname\ can learn symbolic functions by applying the same policy to select their outputs. The key contributions of \shortname\ are:

\begin{itemize}
    \item \emph{Learning the symbolic and the perception function through supervision only on the NeSy function}. To the best of our knowledge, \shortname\ is the first NeSy system that can simultaneously learn symbolic and perception functions in an end-to-end fashion, from supervision only on their composition and with minimal biases on the symbolic function. It has been shown that previous such claims \cite{wang2019satnet} contained some form of label leakage leading to supervision on the individual perception functions \cite{SAT_NET_problem}, and the system completely fails (with 0\% accuracy on visual-sudoku task) when supervision on the perception function is removed. Furthermore, later works on this idea rely on clustering-based pre-processing \cite{SAT_Net_Solve} and do not constitute an end-to-end system. 

    \item \emph{Symbol Grounding Problem} (SGP) refers to the problem of associating symbols to abstract concepts without explicit supervision \cite{Symbol_Grounding_Problem} on this association. The SGP is considered a major prerequisite for intelligent agents to perform real-world logical reasoning. Recent works \cite{SAT_NET_problem} have also provided extensive empirical evidence on the non-triviality of this task, even on the simplest of problems. In \shortname\, we can create internal (interpretable) symbolic representations that are then associated to perception inputs (e.g., handwritten digits) while getting supervision only on higher order operations (e.g., the sum of the digits). Furthermore, unlike previous works \cite{SAT_Net_Solve}, DSL does not rely on clustering based pre-processing. This is important as such pre-processing informs the system about the number of symbols, whereas DSL can infer such number and create meaningful associations between symbols and perception inputs.

    \item  \emph{Differentiable Discrete Choices.} \shortname\ is the first NeSy architecture that provides a method for making discrete symbolic choices within an end-to-end differentiable architecture. It achieves this by exploiting a \emph{policy function} that, given confidence values on an arbitrarily large set of symbols, is able to discretely choose one of them. Furthermore, the policy function can be changed to exploit varying strategies for the choice of symbols.
    
\end{itemize}
Finally, we provide extensive empirical verification of the aforementioned claims by testing DSL on three different tasks.
Firstly, we test our system on a variant of the MNIST \cite{lecun1998gradient} Sum task proposed in \cite{manhaeve2018deepproblog}, where the knowledge about the sum operation is not given but learned (see Example \ref{ex: MNIST_SUM}).
Moreover, we also test \shortname\ with no prior information on the number of required internal symbols, showing that \shortname\
can correctly associate them with perception inputs while learning the summation rules. DSL provides competitive results, even in comparison to systems that exploit prior knowledge. 

Finally, in the last two experiments, we test a recursive variant of \shortname\ on the Visual Parity (see Example \ref{ex: parity}) and the Multi-digit Sum tasks (see Example \ref{ex:multi_digit}). In these experiments, DSL shows great generalization capabilities. Indeed, we trained it on short sequences, finding that it can generalize to sequences of any length.

%% file: chapters/related_work.tex
NeSy has emerged as an increasingly exciting field of AI, with several directions \cite{NeSy_Survey}. Approaches like  Logic Tensor Networks \cite{LTN} and Semantic-Based Regularization \cite{SBR} encode logical knowledge into a differentiable function based on fuzzy logic semantics, which is then used as a regularization in the loss function. Semantic Loss \cite{SL}, also aims at guiding the NN training through a logic-based differentiable regularization function based on probabilistic semantics, which is obtained by compiling logical knowledge into a Sentential Decision Diagram (SDD) \cite{SDD}. In comparison to \shortname\, these approaches assume that the symbolic function is already given and is not learned from data. Furthermore, the symbolic function is only used to guide the learning of the perception function and does not influence the NN predictions at test time.

A parallel set of approaches incorporates NN's as atomic inputs to the conventional symbolic solvers. DeepProbLog \cite{manhaeve2018deepproblog}, a neural extension to ProbLog \cite{ProbLog}, admits neural predicates that provide the output of an NN, interpreted as probabilities. The system then exploits SDDs enriched with gradient semirings to provide an end-to-end differentiable system for learning the NN and the program parameters simultaneously. Recent works have aimed at providing similar neural extensions to other symbolic solvers. DeepStochLog \cite{winters2022deepstochlog} and NeurASP \cite{yang2020neurasp} provide such extensions to Stochastic Definite Clause Grammars and Answer Set Programming respectively. In comparison to the regularization-based approaches, these approaches are able to exploit the symbolic function at inference time. However, they also assume the symbolic function to be given. NeSy methods like NeuroLog \cite{tsamoura2021neural}, ABL \cite{ABL} and ABLSim \cite{ABL_Sim} are based on abduction-based learning framework, where the perception functions have supervision on assigning symbolic labels to perception data. However, the reasoning framework provides additional supervision to make the perception output consistent with the knowledge base i.e., the symbolic function. An abduction based  approach closely related to our work is MetaAbd \cite{Meta} where latent symbols are associated to perception inputs, while simultaneously learning a logical theory and  latent symbols based on a probability-based score function. However, in comparison to DSL, MetaAbd provides complex logical primitives, whereas DSL can learn logical rules with minimal prior bias. Furthermore, MetaAbd samples the space of logical hypothesis, whereas DSL is an E2E differentiable framework learning logical theories and perception symbols within the same differentiable pipeline. Apperception Engine \cite{Evans} also aims at learning logical theories from raw data. However, when raw data is continuous, i.e., consists of a perception-based tasks like recognizing images, they use pre-trained NNs. Hence, unlike DSL, Apperception Engine cannot simultaneously learn to create symbols for perception inputs and learn logical theories on those symbols. 

Another paradigm of NeSy integration consists of works that aim at learning the symbolic function, with either no perception component or with supervision on the perception function. Neural Theorem Prover \cite{NTP} uses soft unification to learn symbol embeddings to correctly satisfy logical queries. Logical Neural Networks \cite{Logical_Neural_Networks} is a NeSy system that creates a 1-to-1 mapping between neurons and elements of a logical formulae. Hence, treating the entire architecture as a weighted real-valued logic formula. SATNet \cite{wang2019satnet} aims to learn both the symbolic and perception functions. It does so by encoding MAXSAT in a semi-definite programming based continuous relaxation, and integrating it into a larger deep learning system. However, it has been shown that it can only learn the symbolic function when supervision on perception is given \cite{SAT_NET_problem}. \cite{SAT_Net_Solve} extends SATNet to learn perception and symbolic functions, aiming at resolving the symbol grounding problem in SATNet. This extension relies on a pre-processing pipeline that uses InfoGAN \cite{InfoGAN} based latent space clustering. Besides not being end-to-end, their method assumes that the number of symbols (i.e., the number of digits in their experiments) is given apriori. \cite{EMB2SYM} is another approach that exploits latent space clustering for extracting symbolic concepts. Furthermore, they assume the number of symbols and the logical rules to be given apriori. In DSL, only an upper bound needs to be provided on the number of required symbols. If the amount of symbols provided is higher than the correct one, it learns to ignore the additional symbols, mapping the perceptions to only the required number of symbols.

Most of the NeSy systems in literature have distinct perception and symbolic components. To the best of our knowledge, none of these systems can learn both the components from supervision provided only on their composition. In \shortname, we provide an approach that is able to learn both the symbolic functions and the perception functions separately, from supervision only on their composition. Furthermore, the symbols required to create the rules are created internally and are associated to perception within a unique NN learning pipeline. In comparison to the SOTA NeSy methods, \shortname\ is the first end-to-end NeSy system to resolve a non-trivial instance of both the symbol grounding problem and rule learning from perception.

%% file: chapters/problem_definition.tex
\section{Background}
\paragraph{Notation.}We denote sets with math calligraphic font and its elements with corresponding indexed lower case letter, e.g., $\Sym = \{ s_i | \forall i \in \mathbb{N}, 0 < i \leq k \}$, where $k = |\Sym|$ is the cardinality of the set. 
Tensors are denoted with capital bold letters (e.g. $\G$) and the $[ \dots ]$ operator is used to index values in a tensor. For instance, given a matrix $\G \in \mathbb{R}^{10 \times 10}$, the element $\G[1, 2]$ corresponds to the entry in row 1 and column 2 of $\G$. Similar to python syntax, we introduce the colon symbol for indexing slices of a tensor. As an example, $\G[1, \ :]$ corresponds to the vector $\langle \G[1, 1], \dots, \G[1, 10] \rangle$. We use a bar on top of functions and elements of a set to denote a tuple of functions and elements, respectively. For instance, $\bar{s} = \bar{f}(\bar{x})$ is equivalent to:
$$(s_1, \dots, s_n) = (f_1(x_1), \dots, f_n(x_n))$$
Note that the length $n$ of the tuple is omitted from the bar notation since it will always be clear from the context.  

\paragraph{Fuzzy Logic.}
Fuzzy Logic is a multi-valued generalization of classical logic, where truth values are reals in the range $[0,1]$. In this work, we will only be dealing with conjunctions, which in fuzzy logic are interpreted using \emph{t-norms}. A t-norm $t:[0,1]\times[0,1]\rightarrow [0,1]$ is a function that,  given the truth values $t_1$ and $t_2$ for two logical variables, computes the truth value of their conjunction. In this paper, we will exploit \godel\ t-norm, which defines the truth value of a conjunction as the minimum of $t_1$ and $t_2$.

\paragraph{Problem Definition.}
\begin{figure*}
\begin{center}
    \includegraphics[scale=0.1]{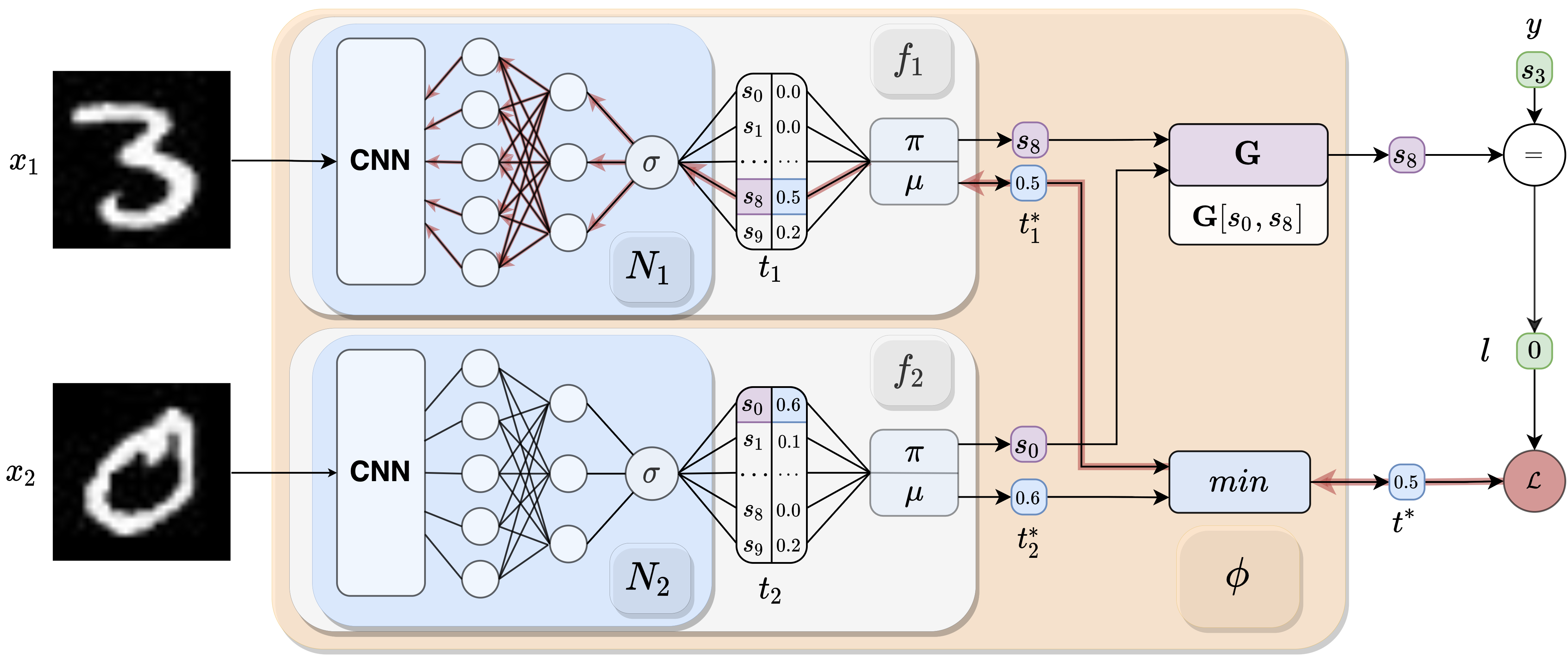}
\end{center}
\caption{\label{fig:architecture} Architecture of \name\ for the Sum task. Red arrows represent the backward signal during learning.} 
\end{figure*}

{ Our approach to NeSy can be abstractly described as the problem of jointly learning a set of perception and symbolic functions providing supervision only on their composition. We define $\mathcal{X}$ to be the space of 
possible perception inputs. Given a finite set $\Sym$ of discrete symbols, a \emph{perception functions}
$f: \mathcal{X}\rightarrow \Sym$ maps from $\mathcal{X}$ to symbolic output in $\Sym$. We define a  \emph{symbolic function}, $g:\Sym^{n} \rightarrow \Sym$, that maps an $n$-tuple of symbols to a single output symbol. We will also consider $g$ with a typed domain, i.e., given some sets of symbols $\Sym_1,...,\Sym_n$ and $\Sym$, $g$ could map from $\Sym_1 \times \cdots \times \Sym_n$ to $\Sym$. Finally, we define a \emph{NeSy functions} $\phi: \mathcal{X}^{n}\rightarrow \Sym$ as a composition of perception and symbolic functions. 
In this paper, we will provide supervision only on the NeSy-function through a training set $Tr$ of the form  
$
Tr = \bigl\{ \bigl( \bar{{x}}_i, y_i \bigr) \bigr\}_{i = 1}^{m}
$, 
where $y_i = \phi(\bar{x}_i)$ and $m$ is the dimension of the training set.
The goal is learning both the NeSy function and its components.
NeSy-functions can constitute arbitrary compositions of symbolic and perception functions. In this paper we consider two such cases, namely  \emph{Direct NeSy function}, and \emph{Recurrent NeSy function}.}

\begin{definition}[Direct NeSy function]
Let $g:\Sym_1\times\dots\times \Sym_n\rightarrow \Sym$ be a symbolic function and $f_i:\mathcal{X}\rightarrow \Sym_i$, for $i=1,\dots,n$ be 
$n$ perception functions. A \emph{Direct NeSy-function} is defined as the composition of $g$ with the $f_i$
\begin{align}
\phi( x_1,\dots, x_n) & = g(f_1( x_1),\dots,f_n( x_n))
\end{align}
\end{definition}

\begin{example}[Sum task]
\label{ex: MNIST_SUM}
Let $\Sym_1$ and $\Sym$ be the following set of symbols: 
  $\Sym_1$ are the integers from $0$ to $9$ and $\Sym$ is the set of integers from $0$ to
  $18$. Let us have a training set, consisting of tuples $(x_1,
  x_2 ,y)$ where $x_1$ and $x_2$ are images of handwritten
  digits  and
  $y$ is the result of adding the digits $x_1$ and $x_2$. Our goal is to learn the Direct NeSy-function:
  $$
  \phi(x_1, x_2) = g(f(x_1),f(x_2))
  $$
  where $f$ is handwritten digit classifiers. Hence, our goal is to learn $f$ and $g$ with supervision provided only on $g(\bar{f}(\bar{x}))$.
\end{example}

As a second type of composition we will consider Recurrent NeSy functions, i.e., NeSy functions defined recursively. In general, it is possible to define complex types of recurrent compositions involving multiple perception and symbolic functions. In this work, we focus on a few such possibilities.

\begin{definition}[{Simple} Recurrent NeSy-function]
Let $g: \Sym \times \Sym \rightarrow \Sym$ be a symbolic function and $f:\mathcal{X} \rightarrow \Sym$ be a perception function. Moreover, it is given an ordered list of perceptions 
$X = \{ x^{(k)} \}_{k=1}^{K}$, with $x^{(k)} \in \mathcal{X}$. We define $X^{(k)}$ as the sequence of first $k$ elements of $X$.
A \emph{{Simple} Recurrent NeSy-function} $\phi$ is defined recursively as:
\begin{align*}
&\phi(X^{(k)}) = g(f(x^{(k)}), \phi(X^{(k-1)}))\\
&\phi(X^{(0)}) = s^{(0)} \in \Sym
\end{align*}
\label{eq:simple_recurrent}
\end{definition}

\begin{example}[Visual Parity]
\label{ex: parity}
Let $\Sym = \{s_0, s_1\}$ be a set composed of two symbols, representing binary values, and $\phi(X)$ the {Simple} Recurrent NeSy function which represents the parity function, i.e., the function that returns $s_0$ if the number of $s_1$ in the sequence is even, $s_1$ if it is odd. $\phi(X)$ can be expressed in terms of a perception function $f$ and a symbolic function $g$ using previous definition of {Simple} Recurrent NeSy function: the $f$ converts the perceptions in binary values, while the $g$ represents the XOR operator, with $s^{(0)} = s_0$.

\end{example}

\begin{example}[Multi-digit Sum]
\label{ex:multi_digit}
Let $\Sym$ be the set of symbols corresponding to the integers from 0 to 9, and $\Sym_c = \{s_0, s_1\}$ another set of symbols. We have a training set composed of pairs of multi-digit numbers and their sum as labels. Each number is represented by a list of MNIST digit images. The goal is to learn the NeSy function $\phi$ that computes the sum of the given numbers. Similarly to Example~\ref{ex: parity}, the NeSy function can be defined recursively. However, in this case, there are two symbolic functions, which compute the single-digit summation modulo 10 and the carry, respectively. For more details, we refer the reader to the Supplementary Material.
\end{example}

%% file: chapters/method.tex
\paragraph{Policy Functions.}

In this paper we will exploit the concept of \emph{policy functions} inspired by Reinforcement Learning (RL). In RL, an agent has at its disposal a set of available actions, and at each time frame only one action can be performed. The goal is to select actions that maximize the expected reward. A strategy for choosing the actions, based on the current state of the system, is called a \emph{policy}. In this work we consider two specific policies, namely the \emph{greedy} and the $\emph{$\epsilon$-greedy}$, and we adapt them to the context of NeSy. In our setting, a \emph{policy} selects a symbol instead of an action, and it is defined as a function $\pi: [0,1]^{|\Sym|} \rightarrow \Sym $ that, given a vector $t \in [0,1]^{|\Sym|}$, returns a symbol $s_i \in \Sym$. Intuitively, $t$ is a vector of confidences returned by a neural network, which in our framework are interpreted as a vector of fuzzy truth values. Formally, $t_i$ corresponds to the truth value of the proposition $(s_i = s^*)$, where $s^*$ is the correct (unknown) symbol. Moreover, we define
the function $\mu: [0,1]^{|\Sym|} \to [0,1]$ as the function that returns the truth value of the symbol chosen by the policy.

The \emph{greedy policy} selects the symbol with highest truth value: $\pi(t) = argmax_i t_i$. The function $\mu$ returns the corresponding truth value: $\mu(t) = max_i t_i$. \shortname\ exploits the differentiability of $\mu$ to indirectly influence the policy $\pi$, which is not differentiable. In the case of the greedy policy, by decreasing the highest confidence ($\mu(t)$), we reduce the chances for the current symbol to be selected again.

\emph{$\epsilon$-greedy} behaves like the greedy policy with probability $1 - \epsilon$, while it chooses a random symbol with probability $\epsilon$. The advantage of $\epsilon$-greedy over greedy is a better ability to explore the solutions space. In our experiments, we use $\epsilon$-greedy during training, and greedy policy at test time.

\paragraph{\shortname\ for Direct NeSy-functions.}
For sake of presentation, we first assume symbolic functions to be given, and our goal is to learn the perception functions. We will then extend \shortname\ to learn also the symbolic function.

We first define 
the representation of the perception functions $f_i: \mathcal{X}\rightarrow \Sym_i$
  and the symbolic function $g$. 
  W.l.o.g., we assume that symbols in
  any set $\Sym_i$ are represented by integers from 1 to $|\Sym_i|$. The
  symbolic function $g:\Sym_1 \times \cdots \times \Sym_n \rightarrow \Sym$ is 
  stored as a $|\Sym_1| \times \cdots \times |\Sym_n|$ tensor
  $\G$, where $\G[s_1,\dots, s_n]$ contains the integer representing the
  symbolic output of $g(\bar{s})$. Every perception function
  $f_i: \mathcal{X}\rightarrow \Sym_i$ is modelled as $\pi(N_i)$, where
  $N_i : \mathcal{X} \rightarrow [0,1]^{|\Sym_i|}$ is a neural network (NN), and 
  $\pi:[0,1]^{|\Sym_i|} \rightarrow \Sym_i $ is a \emph{policy
    function}. For every $x\in \mathcal{X}$, $N_i(x)$ is an
  $|\Sym_i|$-dimensional vector $\bar{t}_i\in[0,1]^{|\Sym_i|}$ 
  whose entries sum to 1. Intuitively, the
  $l^{th}$ entry in $\bar{t}_i$ represents the predicted truth value associated with the $l^{th}$
  symbol being the output of $f_i(x)$. The policy function $\pi$
  makes a choice and picks a single symbol from $\Sym_i$ based on
  $\bar{t}_i$. In summary, our model is defined as:
$$
\phi'(\bar{x}) = \G[\pi(N_1(x_1)), \dots, \pi(N_n(x_n))]
$$
where $\phi'$ is the learned approximation of target function $\phi$.

\begin{example}[Example 1 continued]
\label{ex: MNIST_SUM_Truth}
We assume the same setup as Example \ref{ex: MNIST_SUM}, with an addition that $f_i(x_i)$ is $\pi(N_i(x_i))$ (with $i \in {1,2}$), as presented above. Let the prediction of $f_1$ and $f_2$ be the integers $3$ and $5$ respectively. In this context, $\G$ is a matrix that contains the sum of every possible pair of digits, so that $\G[i,j] = i + j$. Therefore, the prediction is: $\phi'(\bar{x}) = \G[3, 5] = 8$.
\end{example}
\paragraph{Learning the Perception Functions.}
In example~\ref{ex: MNIST_SUM_Truth},
if one of the two internal predictions were wrong, then the final prediction $8$ would be wrong as well. Hence, we define the confidence of the final prediction to be the same as the confidence of having both internal symbols correct simultaneously.
In other words, we could consider the output $\phi'(\bar{x})$ of the model to be correct if the following formula holds for all perception input $x_i$:
\begin{equation}
\bigl(\pi(N_1(x_1)) = s_1^*\bigr) \land \dots \land \bigl(\pi(N_n(x_n)) = s_n^*\bigr)
\label{eq:logic_formula}
\end{equation}
where $s_i^*$ is the (unknown) ground truth symbol associated to perception $x_i$. We interpret formula in Equation~\ref{eq:logic_formula} using \godel\ semantics, where the conjunctions are interpreted by the $min$ function. 
We use $t_i^*$ to denote the truth value (or the confidence) given by the NN for the symbol selected by $\pi$, i.e.,  $t_i^* = \mu(N_i(x_i))$. 
Hence, the truth value $t^*$ associated to the final prediction $\phi'(\bar{x})$ is given as:
\begin{equation}
\label{eq: truth_perception}
    t^* = min_i \ t^*_i = min_i \ \mu(N_i(x_i))
\end{equation}

\begin{figure}
\begin{center}
    \includegraphics[scale=0.12]{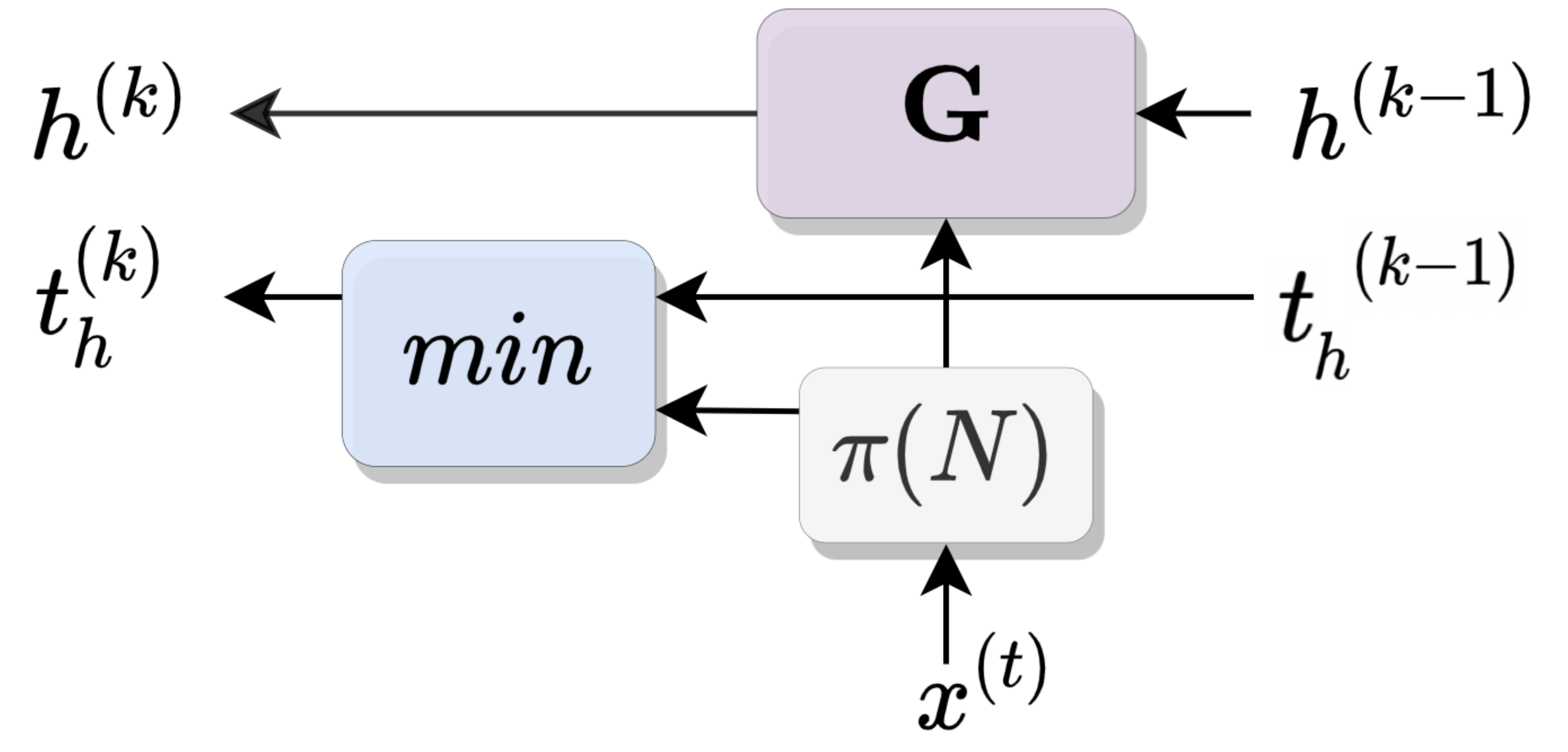}
\end{center}
\caption{\label{fig:architecture_recursive} Architecture of \name\ for the simple recurrent NeSy functions.}
\end{figure}

To train the model we use the binary cross entropy loss on the confidence $t^*$ of the predicted symbol. 
If it is the right prediction, the confidence should be increased. In such a case, the ground truth label is set to one.
If $\phi'(\bar{x})$ is the wrong prediction, the confidence should be reduced, and the label is set to zero. In summary, the entire architecture is trained with the following loss function:
$$
\mathcal{L} = - \sum_{\bm (\bar{x}, y) \in Tr} l \cdot \log(t^*) + (1 - l) \cdot \log(1 - t^*)
$$
where $
l = \mathbbm{1}(\phi'(\bar{x}) = y)
$, and $\mathbbm{1}$ is the indicator function. The architecture is summarized in Figure~\ref{fig:architecture}, where we show an instance of \shortname\ in the context of Example~\ref{ex: MNIST_SUM}.
\paragraph{\shortname\ for Recurrent NeSy-functions.}
In DSL, a simple recurrent NeSy function is represented recursively as:
\begin{align*}
&\phi'(X^{(k)}) = \G[\pi(N(x^{(k)})), \phi'(X^{(k-1)})]\\
&\phi'(X^{(0)}) = \pi(\sigma(\W_0))
\end{align*}
where $\W_0 \in \mathbb{R}^{|\Sym|}$ is the set of weights associated to the initial output symbol. 
Again, we define $t^* = min_i t_i^*$ as the minimum among the truth values of the internally selected symbols. The architecture is presented in Figure~\ref{fig:architecture_recursive}.
It is worth noticing the similarity between the \shortname\ model and the Equation~\ref{eq:simple_recurrent}. In general, a \shortname\ model can be instantiated by following the same compositional structure of the NeSy function we want to learn, applying the policy when a value is expected to be symbolic. For instance, in the multi-digit task of Example~\ref{ex:multi_digit}, we can change the model by exploiting two distinct matrices ($G_c$ and $G_s$) of size $[10, 10, 2]$, instead of one. $G_c$ maps the two current images and the carry to two possible outputs (the next carry values), while $G_s$ to 10 (the output digits). Differently from the visual parity case, here the output is a list of numbers, whose dimension is same as the inputs (e.g., [3,2]+[4,1]=[7,3]) or one digit longer (e.g., [9,2]+[4,1]=[1,3,3]).
For this reason, we add a padding consisting of zeros at the beginning of the input lists, making their length the same as the output.

More details about the architecture for the multi-digit sum can be found in the Supplementary materials.

\paragraph{Learning Symbolic Functions.}
So far we have assumed the symbolic function $g$ to be given. We now lift this assumption and define a strategy for learning the $g$. The idea comes from a simple observation: for each tuple $\bar{s}$ there exists exactly one output symbol $g(\bar{s})$. Note that the mechanism introduced to select a unique symbol from the NN output can be also used for selecting \emph{propositional symbols}, i.e. static symbols that do not depend on the current perceptions. We use the policy functions on learnable weights, allowing to learn the symbolic rules directly from the data.

\begin{figure}[t]
\begin{center}
    \includegraphics[scale=0.1]{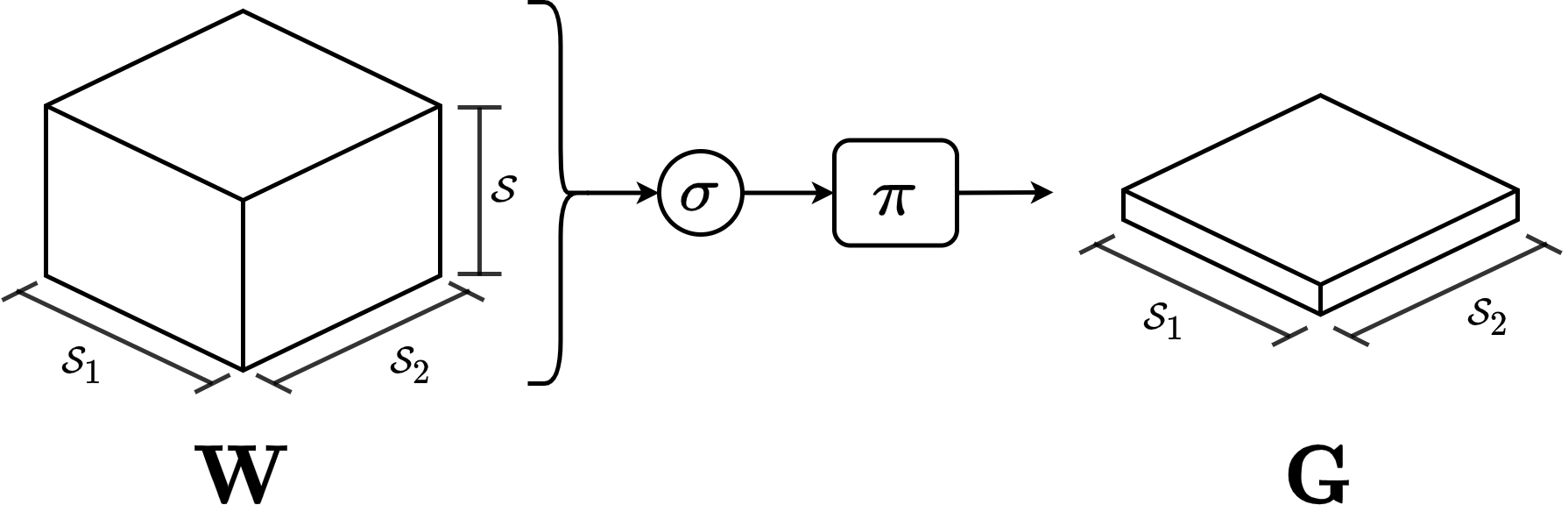}
\end{center}
\caption{\label{fig:learning_g} Tensor $\W$ is used by the policy to generate tensor $\G$. This is done by applying the policy on the output dimension (vertical axes in the image), selecting a single output element for each pair of symbols $(s_1, s_2) \in \Sym_1 \times \Sym_2$.
}
\end{figure}

Formally, we define a tensor $\W \in \mathbb{R}^{|\Sym_1| \times \cdots \times |\Sym_n| \times |\Sym|}$ as the weight tensor of $\G$. Note that the tensor shape is the same as $\G$, except for the additional final dimension, which is used to store the weights for all of the output symbols. The entry in $\G$ corresponding to tuple $\bar{s}$ is defined as:
$$
\G[\bar{s}] = \pi(\sigma(\W[\bar{s}, \ :]))
$$
where the softmax function $\sigma$ and the policy $\pi$ are applied along the last dimension of $\W$. The method is summarized by Figure~\ref{fig:learning_g}.

Since the tensor $\G$ is now learned, we need to consider the confidence associated with the choice of symbols in $\G$. The confidence of the final prediction is now defined as 
\begin{equation}
\label{eq: truth_rule}
    t^* = min(t^*_G, min_i \ t_i^*)
\end{equation}
where $t^*_G$ is the confidence of the output symbol for the current prediction:
$$
t^*_G = \mu(\sigma(\W[\bar{s}, \ :])
$$
with $\bar{s} = \pi(\bar{N}(\bar{x}))$ corresponding to the tuple of predictions made by the perception functions.
\paragraph{Gradient Analysis for the Greedy Policy.}
We analyze the partial derivatives of the loss function with respect to the truth values $t^*_i$ and $t^{*}_{G}$. However, notice that $t^{*}$ in equation \eqref{eq: truth_perception} and \eqref{eq: truth_rule} are computed by selecting minimum over $\{t^{*}_i\}^{|S|}_{i=1}$ and  $\{t^{*}_i\}^{|S|}_{i=1} \cup \{t^{*}_{G}\}$ respectively. Hence, for simplicity of notation, in this analysis we denote $t^{*}_{G}$ by $t^{*}_{0}$. We consider only a single training sample, and assume that the policy is the greedy one.
\begin{align*}
\frac{\partial \mathcal{L}}{\partial t^{*}_i} &= - l\frac{\partial \log(t^{*})}{\partial t^{*}_i} - (1-l)\frac{\partial \log(1-t^{*})}{\partial t^{*}_i}\\
&= - \frac{l^{ \ }}{t^{*}}\frac{\partial t^{*}}{\partial t_i^{*}} + \frac{1 - l^{ \ }}{1-t_i^{*}}\frac{\partial t^{*}}{\partial t_i^{*}}
\end{align*}
Now, since $t^{*}$ is the minimum of all $\{t^{*}_i\}^{|S|}_{i=0}$, the term $\frac{\partial t^{*}}{\partial t_i^{*}}$ is 1 if $t_i$ is the minimum value in  $\{t^{*}_i\}^{|S|}_{i=0}$ and 0 otherwise, reducing the total gradient to the following equation :
\begin{equation}
\label{eq: gradient}
 \frac{\partial \mathcal{L}}{\partial t^{*}_i} = \begin{cases}
- \frac{l^{ \ }}{t^{*}} + \frac{1-l^{ \ }}{1-t^{*}} & i = \argmin_{j} t^{*}_j\\
0 &\text{otherwise}
\end{cases}
\end{equation}

For each sample, only one confidence value $t^*_i$ has a non-zero  gradient, meaning that only a single symbolic choice is supervised, i.e., 
the choice of a rule symbol (when $i=0$ in equation \eqref{eq: gradient}) or the choice of a perception symbol. Hence, depending on the performance on a given sample, \shortname\ manages to modify either a perception or the symbolic function. This behaviour is shown in Figure~\ref{fig:architecture} by using red arrows to represent the backward signal generated by the backpropagation algorithm. The signal moves from the loss to $f_1$, which corresponds to the symbol with lower confidence, and when it reaches the softmax function ($\sigma$), it is spread to the entire network. In \shortname, we have not only interpretable predictions, but gradients are interpretable as well. Indeed, for each sample, there is a unique function $f_i$ or $g$ taking all the blame (or glory) for a bad (or good) prediction of the entire model $\phi'$.

%% file: chapters/experiments.tex
We evaluate our approach on a set of tasks where a combination of perception and reasoning is essential. Our goal is to demonstrate that: $i)$ \shortname\ can learn the NeSy function while simultaneously learning the two components $f$ and $g$, in an end-to-end fashion (\hyperref[chap:MNIST-SUM]{MNIST sum}); $ii$) The perception functions $f_i$ learned on a given task are easily transferable to new problems, where the symbolic function $g$ has to be learned from scratch, with only a few examples (\hyperref[chap:MNIST-MIN]{MNIST Minus - One-Shot Transfer}); 
$iii$) \shortname\ can also be generalized to problems with a recurrent nature \hyperref[chap:MNIST-PAR]{MNIST visual parity}, $iv$) when we provide a smaller representation for $G$ 
\shortname\ can solve harder tasks, like the multi-digits sum. Furthermore, it can easily generalize up to $N$-digits sum, with a very large $N$ (\hyperref[chap:MultiDig]{MNIST Multi-Digit sum}).
\paragraph{Evaluation.}

\begin{figure}[t]
\begin{center}
    \includegraphics[scale=0.15]{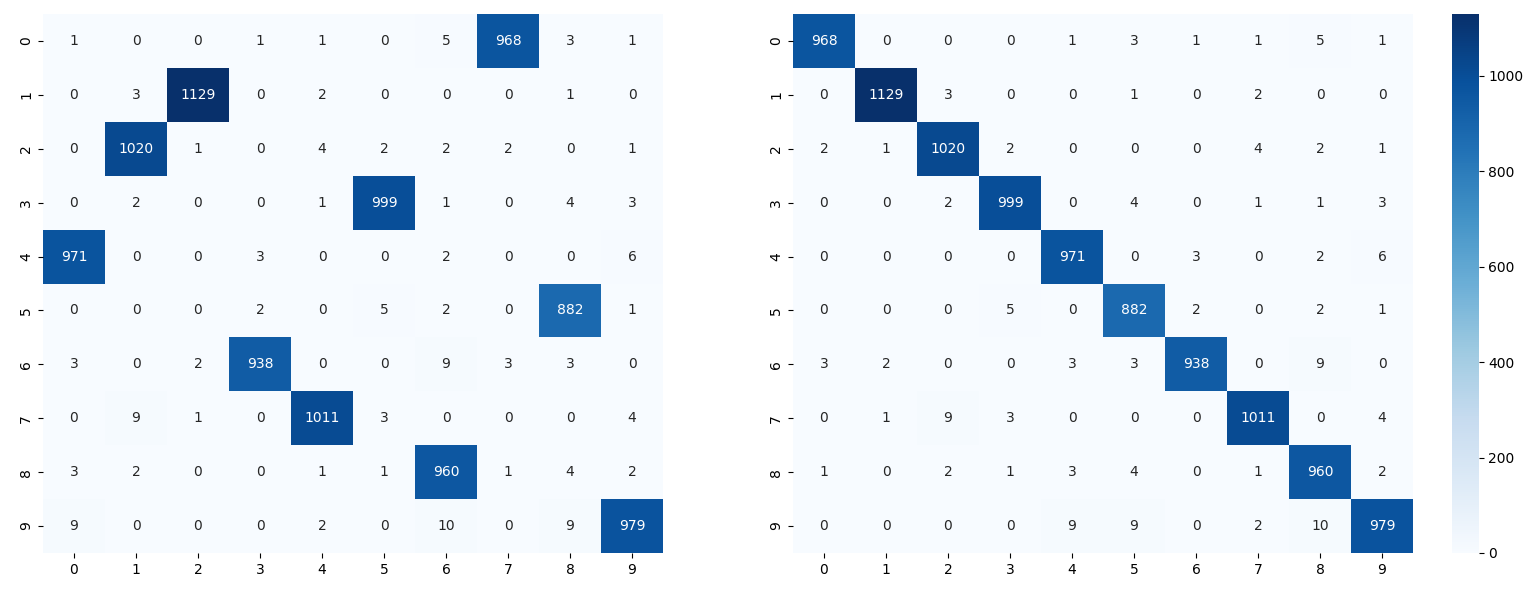}
\end{center}
\caption{\label{fig:confusion} Confusion matrix for the MNIST digits: (left) before the permutation; (right) after permutation.}
\end{figure}

The standard metric used for this type of tasks is the accuracy applied directly to the predictions of the NeSy function $\phi$. This allows us to understand the general behavior of the entire model. However, different from previous models, the symbolic function is also learned from the data. For this reason, we also considered the quality of the learned symbolic rules. However, the symbols associated with perception inputs in DSL are internally generated and form a permutation-invariant representation. Any permutation of the symbols leads to the same behavior of the model, given that the same permutation is applied to the indices of the tensor $\G$. 
Hence, to evaluate the model on learning of $g$s, we need to select a permutation that best explains the model w.r.t the \say{human} interpretation of symbols for digits. The problem is highlighted in Figure~\ref{fig:confusion}(left), where the confusion matrix of the MNIST digit classifier is introduced. Note that for each row (digit), only one column (predicted symbol) has a high value. The same is true for the columns. The network can distinguish the various digits, but the internal symbols are randomly assigned. 
To obviate this problem, we calculate the permutation of columns of the confusion matrix which produces the highest diagonal values (Figure~\ref{fig:confusion}(right)). We then apply the same permutation on the confusion matrix and $\G$, allowing us to obtain a human interpretable set of rules. For more details on this procedure, see Supplementary materials.



\paragraph{Implementation Details.}
All the experiments were conducted with a machine equipped with an NVIDIA GTX 3070 with 12GB RAM. For digit classification, we use the same CNN as \cite{manhaeve2018deepproblog}. We used MadGrad~\cite{defazio2022adaptivity} for optimization and optuna to select the best hyperparameters for every experiment. Results are averaged over 10 runs. 
\paragraph{MNIST Sum.}
\label{chap:MNIST-SUM}
We first tackled the MNIST sum task presented in Example~\ref{ex: MNIST_SUM}. A dataset  consisting of triples $(X, Y, Z)$ is given, where $X$ and $Y$ are two images of hand-written digits, while $Z$ is the result of the sum of the two digits, e.g., ($\digit{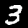}$, $\digit{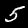}$, $8$). The goal is to learn an image classifier for the digits and the function $g$ which maps digits to their sum.
We implemented two different variants of our approach: \textbf{\shortname} is the naive version of \shortname, where the two digits are mapped to symbols by the same perception function, and the correct number of digits is given a priori; \textbf{\shortname-NB} is a version of \textbf{\shortname} where we removed the two aforementioned biases: we use two different neural networks, $N_1$ and $N_2$, to map perceptions to symbols, and the model is unaware of the right amount of latent symbols, with the neural network returning confidence on 20 symbols instead of 10.
\begin{table}[]
\label{tab:sum}
\centering
    \begin{tabular}{l|l|l}
                                             & Accuracy (\%)    & TE/\#E      \\ \hline
        NAP           & $97.3\pm0.3$    & $109s$/$1$    \\ 
        DPL          & $97.2\pm0.5$    & $367s$/$1$             \\ 
        DStL         & $97.9\pm0.1$    & $25.49s$/$2$             \\ \hline
        \shortname   & $98.8\pm0.3$    & $0.95s$/$50$  \\ 
        \shortname-NB & $97.9\pm0.3$    & $0.99s$/$200$ \\ \hline
    \end{tabular}
    \caption{Results obtained on the MNIST sum task. TE is the time required for 1 epoch, and $\#$E is the number of epochs of training. The SOTA methods are NeurASP (NAP), DeepProbLog (DPL), and DeepStochLog (DStL).}
\end{table}
In table 1, we show that \shortname\ variants have competitive performance w.r.t the state of the art \cite{yang2020neurasp}, \cite{winters2022deepstochlog}, \cite{manhaeve2018deepproblog}. Notice that all the SOTA methods receive a complete knowledge of the symbolic function $g$, while \shortname\ needs to learn it, making the task much harder. Another important result is the accuracy of the DSL-NB method, which proves that \shortname\ can work even with two perception networks and, most importantly, without knowing the right amount of internal symbols. 

\paragraph{MNIST Minus - One-Shot Transfer.}
\label{chap:MNIST-MIN}

One of the main advantages of NeSy frameworks is that the perception functions learned in the presence of a given knowledge ($g$ in our framework) can be applied to different tasks without retraining, just by changing the knowledge. For instance, after learning to recognize digits from supervision on the addition task, methods like DeepProblog can be used to predict the difference between two numbers. However, it is required for a human to create different knowledge bases for the two tasks. In our framework, the $g$ function is learnable, and the mapping from perception to symbols does not follow human intuition (see Evaluation Metrics section). Instead of writing a new knowledge for the Minus task, we replace the tensor $\G$ with a new one and learn it from scratch. In our experiment, we started from the perception function learned from the Sum task and used a single sample for each pair of digits to learn the new $\G$. We obtained an accuracy of $98.1\pm0.5$ after $300$ epochs, each requiring $0.004s$. Note that we did not need to freeze the weights of the $f$. Since the perception functions already produce outputs with high confidence, \shortname\ applies changes mainly on the tensor $\G$.

\paragraph{MNIST Visual Parity.}
\label{chap:MNIST-PAR}

We used the model in Figure~\ref{fig:architecture_recursive} for the parity task using images of zeros and ones from the MNIST dataset, and the same CNN used for the MNIST sum task (with 2 output symbols instead of 10). Learning the parity function from sequences of bits is a hard problem for neural networks, which struggle to generalize to long sequences~\cite{shalev2017failures}. The parity function corresponds to the symbolic function $g$, and learning the perception function is an additional sub-task. We used sequences of 4 images during training and 20 on the test and only provided supervision on the final output. \shortname\ reached an accuracy of $98.7 \pm 0.4$ in 1000 epochs, showing great generalization capabilities. As in other tasks, \shortname\ learned the XOR function perfectly. Tnhe errors made by the model only depend on the perception functions. If the perceptions are correctly recognized, the model works regardless of the sequence length.

\paragraph{MNIST Multi-digit Sum.}
\label{chap:MultiDig}

\begin{table}[]
\begin{tabular}{l|llll}

\multicolumn{5}{c}{Accuracy (\%)}                                                                                       \\

     & \multicolumn{1}{c}{2}              & \multicolumn{1}{c}{4}               & \multicolumn{1}{c}{15}                                                       & \multicolumn{1}{c}{1000}          \\
\hline
NAP  & 93.9 $\pm$ 0.7 & \multicolumn{1}{c}{T/O}             &\multicolumn{1}{c}{T/O}                                                      & \multicolumn{1}{c}{T/O}           \\
DPL  & 95.2 $\pm$ 1.7 & \multicolumn{1}{c}{T/O}             & \multicolumn{1}{c}{T/O}                                                      & \multicolumn{1}{c}{T/O}           \\
DStL & 96.4 $\pm$ 0.1 & 92.7 $\pm$ 0.6  & \multicolumn{1}{c}{T/O}                                                      & \multicolumn{1}{c}{T/O}           \\ \hline
DSL  & 95.0 $\pm$ 0.7 & 88.9 $\pm$ 0.5  & 64.1 $\pm$ 1.5                                           & 0.0 $\pm$ 0.0 \\
\hline
\multicolumn{5}{c}{Fine-grained Accuracy (\%)}\\
     & \multicolumn{1}{c}{2}              & \multicolumn{1}{c}{4}               & \multicolumn{1}{c}{15}                                                       & \multicolumn{1}{c}{1000}          \\
\hline
DSL     & 97.9 $\pm$ 0.1 & 97.3 $\pm$ 0.1 & 96.7 $\pm$ 0.1 & 96.5 $\pm$0.1\\
\hline
\end{tabular}
\caption{Results obtained on the MNIST Multi-digit sum task. T/O stands for timeout.}
\label{tab:md}
\end{table}

The previous experiment on the Visual Parity have demonstrated the ability of \shortname\ to learn recursive NeSy functions. However, this experiment was conducted on a simple task where the number of allowed symbols was limited to two, and the symbolic function $g$ could be directly stored in a 2x2 matrix.
The multi-digit sum is more challenging since the hypothesis space becomes much larger, and we need to learn two symbolic functions ($g_c$ and $g_s$) simultaneously.
Thus, we decided to rely on Curriculum Learning
\cite{curriculum}, where initially we provide only samples composed of a single digit and no padding, reducing the problem to learning the digit sum modulo 10. We then provide another training set composed of two digits numbers and the padding, allowing the model to learn the missing rules. We trained our model on the 2-digits sum and we evaluate the learned model on sequences of varying length, showing the generalization capabilities of DSL.

Table \ref{tab:md} reports the results obtained by NeurASP, DeepProblog, DeepStochLog and DSL. We tested our model on $N$-digit sums, with $N$ up to 1000. Also in this case, \shortname\ learned perfect rules; thus, the accuracy degradation obtained by increasing the value of $N$ is only due to errors made by the perception function ($98.5\%$ accuracy).
For this reason, our performance follows a similar trend of $0.985^{(2N)}$. To better understand the true performance of \shortname, we also measured a fine-grained accuracy that measures the mean ratio of correct digits in the final output. Furthermore, our approach took only 0.27 seconds to infer on the entire test set for $N=1000$, while no other methods scale to more than 4 digits.


%% file: chapters/conclusion.tex
We presented Deep Symbolic Learning, a NeSy framework for learning the composition of perception and symbolic functions. To the best of our knowledge, DSL is the first NeSy system that can create and map symbolic representations to perception while learning the symbolic rules simultaneously. A key technical contribution of DSL is the integration of discrete symbolic choices within an end-to-end differentiable neural architecture. For this, DSL exploits the notion of policy deriving from Reinforcement Learning. Furthermore, DSL can learn the perception and symbolic functions while performing comparably to SOTA NeSy systems, where complete supervision of the symbolic component is given. Moreover, in the multi-digit sum, DSL's inference scales linearly, allowing the evaluation of huge sequences. In the future, we aim to extend DSL to problems with a larger combinatorial search space. To this end, we aim to consider factorized matrix representations for the symbolic function $g$, and its weight matrix $W$. Furthermore, we aim to generalize DSL to more complex perception inputs involving text, audio, and vision.

%% file: chapters/appendix.tex
\section{Multi-Digit Sum}
In this section we present the Example 3 in detail. Let $\Sym$ be the set of symbols corresponding to the integers from 0 to 9, and $\Sym_c = \{s_0, s_1 \}$ another set of symbols representing an internal hidden state. Intuitively, after training, those symbols should represent whether the carry is zero or one. However, note that there is no bias on the hidden state being an actual number between zero and one. On the contrary, DSL learns two different behaviors of the sum depending on whether the hidden state is $s_0$ or $s_1$.

We have a training set composed of pairs of multi-digit numbers and their sum as labels. Each number is represented by a list of MNIST digit images. The goal is to learn the NeSy function that computes the sum of the given numbers. Similarly to Example 2, the NeSy function can be defined recursively. However, in this case, there are two symbolic functions, given as follows:
\begin{align*}
    \phi_{s}(X^{(k)},Y^{(k)}) &= g_{s}\big(f(X^{(k)}),f(Y^{(k)}),\phi_{c}(X^{(k-1)},Y^{(k-1)})\big) \\
    \phi_{c}(X^{(k)},Y^{(k)}) &= g_{c}\big(f(X^{(k)}),f(Y^{(k)}),\phi_{c}(X^{(k-1)},Y^{(k-1)})\big)\\
    \phi_{c}(X^{(0)},Y^{(0)}) &= s^{(0)} \in \Sym_c
\end{align*}

The perception function $f$ maps MNIST images to symbols in $\Sym$.
Both $g_c$ and $g_s$ functions receive as inputs the symbols corresponding to the two current digits and the hidden state (carry) computed in the previous step. 
The symbolic function $g_{s}$ returns the sum modulo 10 of the carry with the two digits, while $g_{c}$ returns the value of the next carry.

Similar to the simple recurrent NeSy function, we can define the DSL model by following the compositional structure of the NeSy function.
In this case, the architecture of \shortname\ is given as follows:

\begin{align*}
    \phi'_{s}(x^{(k)},y^{(k)}) &= \mathbf{G_s}[\hat{x}^{(k)},\hat{y}^{(k)},\phi'_{c}({x}^{(k-1)},{y}^{(k-1)}\big)] \\
    \phi'_{c}(x^{(k)},y^{(k)}) &= \mathbf{G_c}[\hat{x}^{(k)},\hat{y}^{(k)},\phi'_{c}({x}^{(k-1)},{y}^{(k-1)}\big)] \\
    \phi'_{c}(x^{(0)},y^{(0)}) &= \pi(\sigma(\mathbf{W_0}))
\end{align*}
\noindent
where $\hat{x}^{(k)} = \pi(N(x^{(k)}))$, and $\hat{y}^{(k)} = \pi(N(y^{(k)}))$ are the symbolic predictions made by the neural networks for the two current digit images.

\section{MNIST MultiOperation}
In this section, we aim to comprehend the scalability characteristics of \shortname. The MNIST Sum experiment revealed that \shortname\ can successfully complete the task without any bias, however, with a restricted hypothesis space. In this new experiment, we intend to tackle a task with a much larger hypothesis space to evaluate if and how the model can handle more intricate challenges. The task is outlined in the example below.

\begin{example}[MNIST Multiop task]
\label{ex:multiop}
  Let $\Sym_1$, $\Sym_2$, and $\Sym$ be the following set of symbols: 
  $\Sym_1$ are the digits from $0$ to $9$, $\Sym_2$ is the set of symbols with four
  mathematical operations, $\{+,-,\times,\div\}$ and $\Sym$ is the set of integers from $0$ to
  $81$. Let us have a training set, consisting of tuples $(x_1,
  x_2, x_3,y)$ where $x_1$ and $x_3$ are handwritten
  digits, $x_2$ is a handwritten operator and
  $y$ is the result of applying the operator to $x_1$ and $x_3$. Our goal is to learn the Direct NeSy-function:
  $$
  \phi(x_1, x_2, x_3) = g(f_1(x_1),f_2(x_2),f_1(x_3))
  $$
  where $f_1$ are handwritten digit classifiers and $f_2$ is a classifier for
  handwritten mathematical operations. Again, our goal is to learn $f_1$, $f_2$, and $g$ from $g(\bar{f}(\bar{x}))$ with no direct supervision on $g$ and $\bar{f}$.
\end{example}
\label{chap:MNIST-MULTIOP}
We generalize the MNIST Sum task by adding additional operators as perception inputs, as described in Example~\ref{ex:multiop}.
The perceptions are two images from the MNIST dataset and a third symbol representing an operation in $\{+,-,\times,\div\}$. The operation's images are generated from the EMNIST dataset \cite{cohen2017emnist}, from which we extracted the images of letters $A$, $B$, $C$, and $D$, representing $+$, $-$, $\times$, and $\div$ respectively. In order to avoid negative numbers as output or divisions by zero, we take these additional steps: we necessitate in the dataset that for any instance with subtraction operator, i.e., \say{$-$}, the second term of the operation is smaller than the first; \say{$\div$} is interpreted as integer division (e.g. ($5 \div 2 = 2 $)) and it requires the second term to be greater than $0$. Notice that the hypothesis space consists of $81^{400}$ possible symbolic functions $g$. Furthermore, the correct $g$ needs to be identified while simultaneously learning the NNs for digit and operator recognition. 

We trained \shortname\ on this task and obtained an accuracy of $92.2\pm0.4$. We trained our model for $9000$ epochs, with an epoch time of $0.84s$. These results show that \shortname\ can learn complex symbolic functions while simultaneously learning to map multiple perceptions over different domains. Furthermore, this experiment shows that \shortname\ can scale to harder problems, but in such scenario, it requires more training.

\begin{figure*}
\centering
    \begin{subfigure}{0.48\textwidth}
    \centering
        \includegraphics[width=\linewidth]{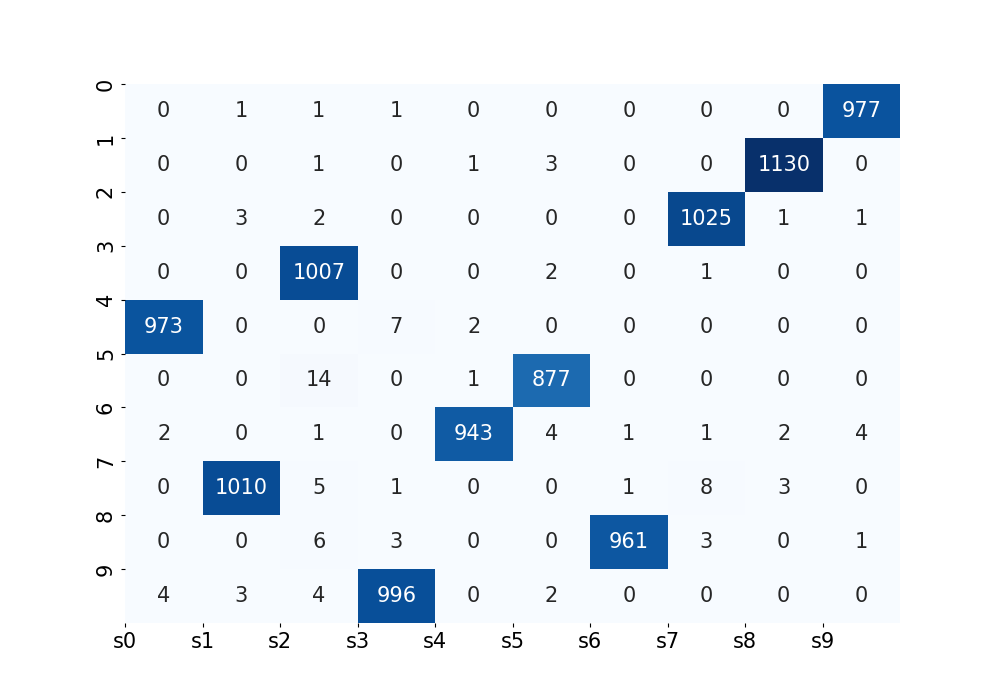}
        \caption{Original Confusion Matrix}
    \end{subfigure}
    \begin{subfigure}{0.48\textwidth}
    \centering
        \includegraphics[width=\linewidth]{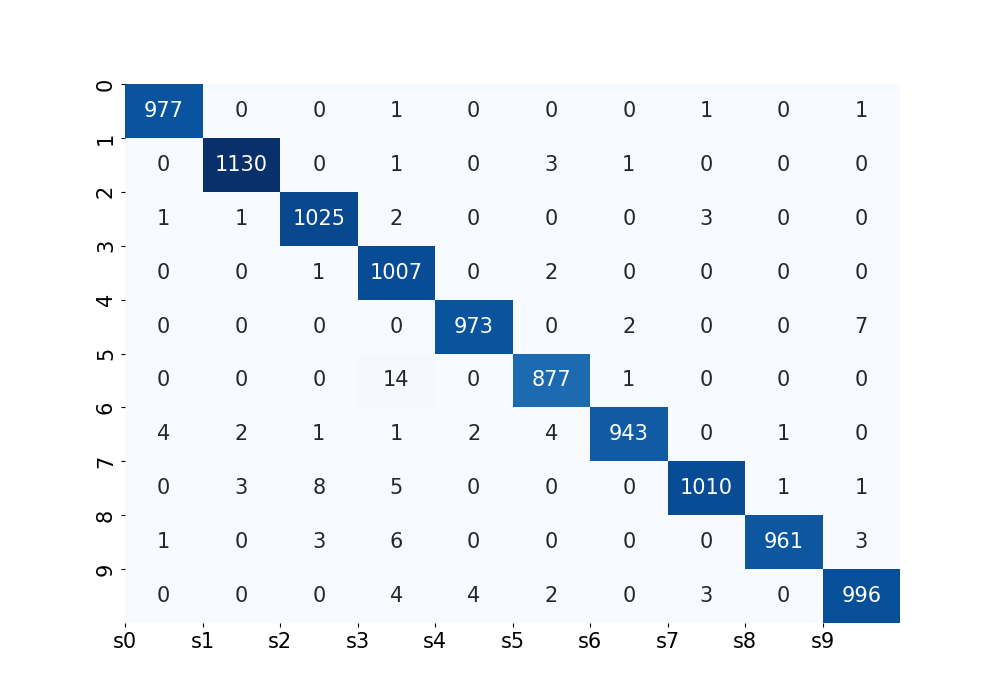}
        \caption{Swapped Confusion Matrix}
    \end{subfigure}
    \begin{subfigure}{0.48\textwidth}
    \centering
        \includegraphics[width=\linewidth]{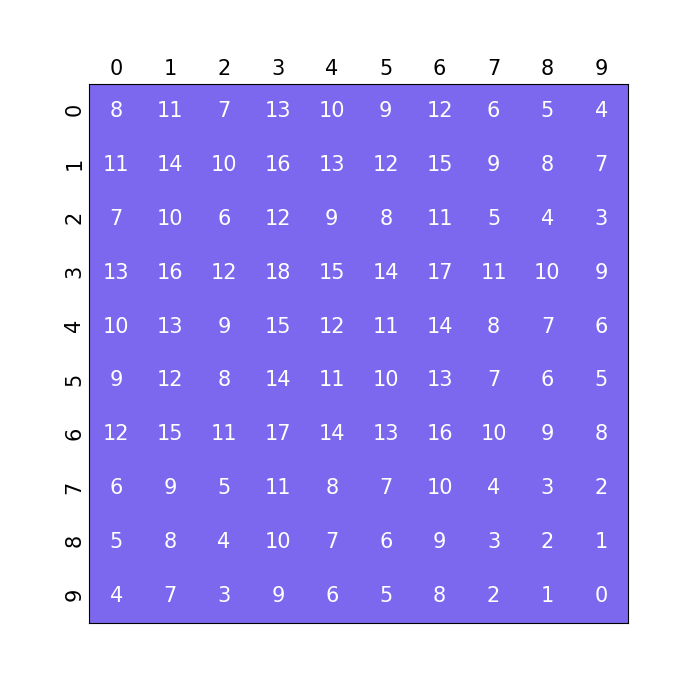}
        \caption{Original Rules Matrix}
    \end{subfigure}
    \begin{subfigure}{0.48\textwidth}
    \centering
        \includegraphics[width=\linewidth]{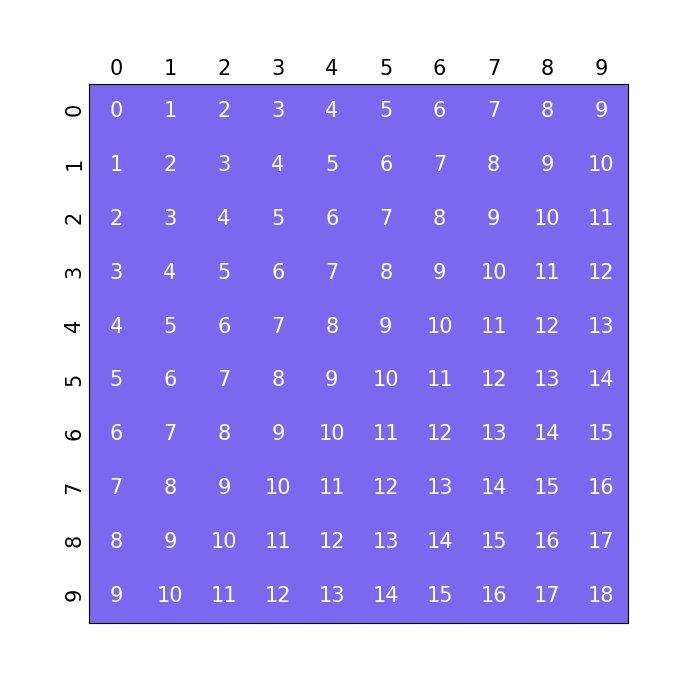}
        \caption{Swapped Rules Matrix}
    \end{subfigure}
    \caption{\shortname\ learns symbolic representations for images, but these representations may not align with our understanding of symbols for digits. To address this, a bijective mapping is applied to connect the DSL symbols to the commonly understood human notion of digits. 
    \shortname\ performs summation correctly in its learned symbolic representation.
    E.g., in the confusion matrix (a), symbol $s_0$ corresponds to digit $4$, and symbol $s_{1}$ corresponds to digit $7$. The learned summation rule for $s_0 + s_1$ can be found in matrix $\mathbf{G}$ (c) in position $(0,1)$, i.e., $\mathbf{G}[0,1] = 11$, which is the correct value for the summation of $4$ and $7$. Evaluating the model's ability to learn symbolic rules becomes easier if we apply the right permutation of the learned symbols.
    As an example, the confusion matrix in (b) is obtained by  permuting the symbols in the x-axis in such a way that the matrix becomes diagonal. The same permutation is applied in (d) to both rows and columns, obtaining a human interpretable rule matrix.}
    \label{fig:swap}
\end{figure*}

\section{Analysis of the symbolic rules matrix}
We introduce here a further explanation of the mechanism used to evaluate the learned symbolic matrix. An example of the procedure is highlighted in Figure~\ref{fig:swap}.